\documentclass[conference]{IEEEtran}

\usepackage[hyphens]{url}
\IEEEoverridecommandlockouts
\usepackage{cite}
\usepackage{amsmath,amssymb,amsfonts}
\usepackage{algorithmic}
\usepackage{graphicx}
\usepackage{textcomp}
\usepackage{xcolor}
\def\BibTeX{{\rm B\kern-.05em{\sc i\kern-.025em b}\kern-.08em
    T\kern-.1667em\lower.7ex\hbox{E}\kern-.125emX}}
\begin{document}

\title{\title{Real-time Emotion and Gender Classification using \\Ensemble CNN}}

\author{\IEEEauthorblockN{ Abhinav Lahariya}
\IEEEauthorblockA{\textit{Indian Institute Of Information Technology ,Allahabad}\\
Prayagraj , Uttar Pradesh , India \\
lahariyaabhinav97@gmail.com}
\and
\IEEEauthorblockN{ Varsha Singh}
\IEEEauthorblockA{\textit{Indian Institute Of Information Technology ,Allahabad}\\
Prayagraj , Uttar Pradesh , India \\
rsi2018002@iiita.ac.in}
\and
\IEEEauthorblockN{ Uma Shanker Tiwar}
\IEEEauthorblockA{\textit{Indian Institute Of Information Technology ,Allahabad}\\
Prayagraj , Uttar Pradesh , India \\
ust@iiita.ac.in}
}

\maketitle

\begin{abstract}
Analyzing expressions on the person’s face plays a very vital role in identifying emotions and behavior of a person. Recognizing these Expressions automatically results in a crucial component of natural human-machine interfaces.Therefore research in this field has a wide range of applications in biometric authentication, surveillance systems,emotions to emoticons in various social media platforms.Another application includes conducting customer satisfaction surveys. As we know that the large corporations made huge investments to get feedback and do surveys but fail to get equitable responses . Emotion \& Gender recognition through facial gestures is a technology that aims to improve product and services performance by monitoring customer behavior to specific products or service staff by their evaluation. In the past few years there have been a wide variety of advances performed in terms of feature extraction mechanisms , detection of face and also expression classification techniques.This paper  is the implementation of an Ensemble CNN for building a real-time system that can detect emotion and gender of the person.The Experimental results shows accuracy of \textbf{68\%} for Emotion classification into 7 classes (angry, fear, sad, happy,surprise, neutral, disgust) on FER-2013 dataset and \textbf{95\%} for Gender Classification (Male or Female) on IMDB dataset.Our work can predict
Emotion and Gender on single face images as well as multiple face images. Also when input is given through webcam our complete pipeline of this real-time system can take less than 0.5 seconds to generate results. 
\end{abstract}

\section{Introduction}
Facial features play a vital role for understanding and recognizing emotions.An important aspect in human interaction is the generality of facial features and gestures. In1971, Friesen and Ekman showed that facial features are universally related with specific emotions. Even animals also depict the same kinds of muscular movements as humans do that belong to a specific state of mind, in spite of their place of race, education, birth, etc. Hence, when modelled properly , this generality can work as a very useful characteristic in human-machine interaction. 

In this work , we try to analyze various facial tasks including Emotion and Gender Classification. Facial images or webcam feed acts as an input to these task so that these tasks predicts their respective classes on the given input. In Emotion classification task , emotion on the person's face classified into seven classes , which are: "angry, disgust, fear, happy, sad, surprise and neutral". Finally we try to predict who are Male or Female among them in Gender classification task.All these tasks can be performed on single faces as well as multiple faces in a single frame.Our complete real time pipeline of face detection , emotion classification and gender classification takes less than 0.5 sec.

FER2013 and IMDB are the two datasets that we used for Emotion and Gender classification task respectively . FER-2013 dataset consists information about 35,887 grayscale, 48x48 sized face images with 7 emotions ,which are labeled as : `` Angry , Disgust , Fear , Happy , Sad , Surprise , Neutral " in which there are 28709 training images and 7178 validation images i.e 80/20 split. IMDB dataset is a very big dataset of faces that contains data of different artists. This dataset consists of approximately 470,000 images.This dataset gives a .mat file as metadata that contains few attributes like face\_score, a second\_face\_score,  gender and age for every image . Images that have  only single frontal face have more face scores, while image containing more than one faces have less face scores. The second\_face\_scores shows how certainly image contains the second face. Further we  take only those  images that contains single face and faces that are mostly frontal. To get this, we select only those images that have face score $\geq 3$. Finally IMDB is splitted into 80/20 ratio for Training and Validation set.

\begin{figure}[ht]
  \centering
  \includegraphics[width=\linewidth]{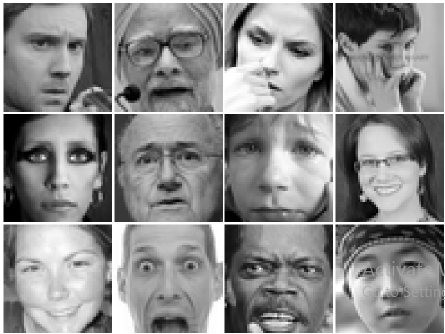}
  \caption{FER-2013 Samples for emotion classification \cite{DBLP:journals/corr/abs-1710-07557} }
\end{figure}

\begin{figure}[ht]
  \centering
  \includegraphics[width=\linewidth]{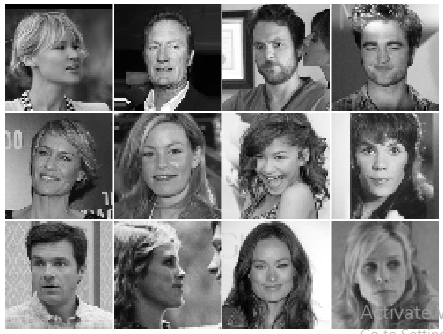}
  \caption{IMDB Samples for gender classification \cite{DBLP:journals/corr/abs-1710-07557} }
\end{figure}

Interpreting Emotion  on the person’s face with the help of Machine Learning(ML) techniques is very complicated due to the large variance in samples from each task. As a result, millions of parameters within the model were trained using thousands of samples. Also, the accuracy of humans in identification of facial expression is $65\% \pm 5\% $.This can be seen by manually classifying the images of  FER-2013 dataset which contains the classes: {“angry”, “disgust”, “fear”,“happy”, “sad”, “surprise”,“neutral”}.

In this paper , we proposed an Ensemble CNN architecture that performs emotion and gender classifcation task and build a real-time system that can take live input from webcam  and predicts emotion and gender of person in a single step.
\section{Related Work }
Large amount of research has been conducted to determine gender and emotion using facial features on different public standard datasets  which permit public performance comparison of the proposed methods .As a result, there has been a lot of active research, with several recent papers using the concept of Convolutional Neural Networks (CNNs) for feature extraction and inference. Facial expressions can be recognised using nonverbal communication between humans, and also facial expression interpretation has been extensively researched. Facial expression is important in human interaction, and the Facial Expression Recognition(FER) algorithm uses computer vision techniques to aid in applications like human-computer interaction and data analytics.

The goal of Xception\cite{DBLP:journals/corr/Chollet16a} Architecture is to provide a powerful method for developing exceptional Deep Learning models. Bigger models are made,either using more depth, that is, adding more sequential layers, or using more number of neurons in different layers of the model. The depth separable convolution can be performed to a great extent since it is connected with the group convolution and the inception modules .The convolutional network \cite{7780677} is located at the center of the highest and the latest and greatest computer vision algorithm and resolution for a wide variety of work. Although the expanded size of model , measurement and data processing costs often provide quality improvements for many tasks, the effectiveness and readiness of the calculation and low parameter requirements still support multi-case components usage. VGG16 is another proposed classifications architecture which can be used  used to recognize emotion, but failed miserably due to its big size and more number of parameters. 

Octavio et.al \cite{DBLP:journals/corr/abs-1710-07557} worked on facial images and proposed a CNN architecture that helps to classify the emotions of facial expression and also classify the gender of a person in a single step. CNN model is implemented with accuracy reaching 96\% for classifying gender in IMDB Dataset and 66\% in classifying emotion in FER-2013 dataset. U.Gogate\cite{9299633} proposed a CNN architecture that helps to classify Emotion of Facial expression with an accuracy of 67\% and also classify the gender of person with accuracy of 95\% in IMDB.Akash Saravanan et.al \cite{DBLP:journals/corr/abs-1910-05602} uses face images of FER2013 dataset to classify the emotions on a person's face into one of the seven categories.In this paper ,a real time emotion detection system is built for emotion classification  with an accuracy reached to 60.58\%.

\section{Methodology}
We are proposing a model architecture that is an ensemble of two CNN models , one is Mini-Xception and another is a simple 4-layer CNN model. We are ensembling these two models using "Average" .Mini-Xception architecture was developed  by FrancoisChollet, who is the creator of Keras library . This is a deep CNN  architecture  that contains depthwise separable convolutions [3].Global average pooling algorithm is used in this model so that it is less dependent on the fully connected layer for trainable parameters.

\begin{figure}[ht]
  \centering
  \includegraphics[width=\linewidth ]{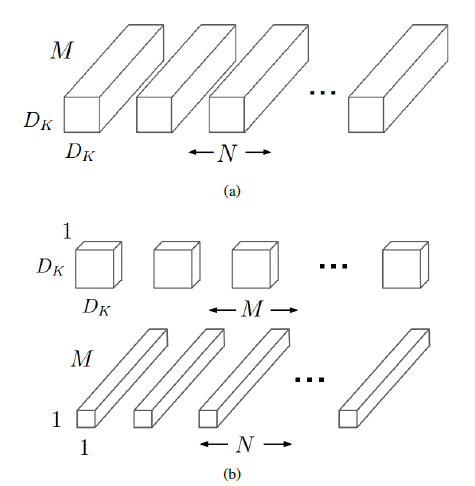}
  \caption{(a) Simple Convolutions (b) Depthwise Separable Convolutions}
\end{figure}

Depthwise Separable Convolutions consist of two types : (1) Depthwise Convolutions (2)  Pointwise Convolutions. The main aim for using these layers is to segregate channel cross-correlations and  spatial cross-correlations. This layer firstly applies a $D \times D$ on each M input channels and after that N $1\times 1\times M$ convolutions filter to integrate M numbers of input channels and N numbers of output channels.This layer minimizes the total computation as compared to normal convolutions by $ 1/N + 1/D^{2} $ \cite{DBLP:journals/corr/HowardZCKWWAA17}

\textit{Why Ensembling ?}
An ensemble is the machine learning model that integrates the predictions of two or more models.Predictions made by different models can be integrated using statistics, such as the mode or mean, or other complex methods. There are mainly two reasons to use an ensemble model over a single model as follows:

\begin{itemize}
    \item \textbf{Performance:}An ensemble model can make better predictions and results in better performance than any single contributing model.
    \item \textbf{Robustness:}An ensemble minimizes the spread of the predictions and model performance.
\end{itemize}

\textit{Why Averaging ?}
Average layer  reduces  the  variance factor in the final neural network model which makes reductions in the spread of the model's performance for getting more confidence in model's result prediction.

\begin{figure}[ht]
  \centering
  \includegraphics[width=\linewidth , height =18cm]{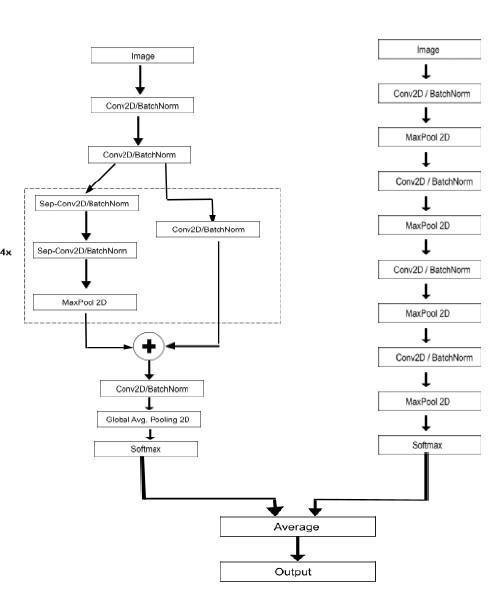}
  \caption{Model Architecture }
  \label{fig:fig4}
\end{figure}

\textbf{Figure \ref{fig:fig4}} shows our proposed model architecture that is our proposed ensembled model.  For emotion classification task we trained our Ensemble model for 100 epochs and for gender classification task we trained only mini-Xception model(part of our ensemble model )for 100 epochs. We used Adam optimizer as an optimization algorithm  and categorical cross entropy Loss as the loss function.We have tested models with proper validation cases  for our emotion and gender classification task on FER-2013 and IMDB dataset respectively. For training purpose we have used Intel(R) Xeon(R) CPU E5-2630 v3 @ 2.40GHz  and 12GB NVIDIA Quadro k6000 GPU.

\section{Experimental Results}
\begin{figure}[ht]
  \centering
  
  \includegraphics[width=\linewidth]{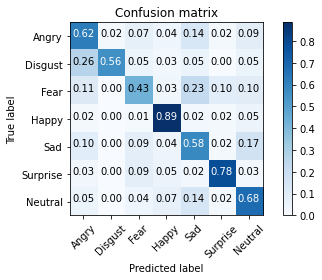}
  \caption{Normalized Confusion Matrix  of our Ensembled model for Emotion classification task on FER-2013 dataset}
  \label{fig:fig5}
\end{figure}

\begin{figure}[ht]
  \centering
  
  \includegraphics[width=\linewidth]{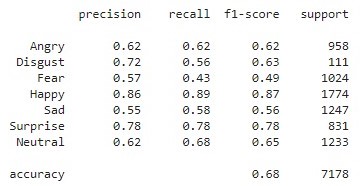}`
  \caption{Results Matrix of our Ensembled model for Emotion classification task on FER-2013 Dataset}
  \label{fig:fig6}
\end{figure}

Results for our real-time emotion and gender classification tasks in hidden faces are noticed.The whole real-time pipeline involves: detection of face , classification  of emotion and Classification of gender  in one single step .This implementation can work on both group as well as single images and also live input given through webcam.

we address our results in \textbf{Table~\ref{tab:table1}} and compare it with other previous approaches .We can see ensemble model gives better performance .This is because we have  used Average layer for ensembling the models to minimize the  variance in the final neural network model which in turn reduces the spread in the model's performance for getting confidence in model's prediction.\textbf{Figure~\ref{fig:fig5}} represents normalized confusion matrix of our Ensembled model for Emotion classification task .\textbf{Figure~\ref{fig:fig6}} shows Precision , Recall , F1- score and support matrices for every class of emotion in Fer-2013 dataset.  For calculating accuracy of the model on FER-2013 and IMDB dataset ,Accuracy metrics is used.The Emotion Recognition task is trained using Ensembled model and achieved accuracy of \textbf{68\%} for test images on FER-2013 dataset.The Gender Classification task uses Mini-Xception model and achieved accuracy of \textbf{95\%}  for test images on IMDB dataset. 
   
\begin{table}{}
\centering
  \caption{Comparsion of Different Approaaches}
  \label{tab:table1}
  \begin{tabular}{ccc}
  \hline\noalign{\smallskip}
    Approaches & FER-2013 & IMDB  \\
     \textbf{Our Approach} & \textbf{68}\% & \textbf{95}\% \\     \textbf{Octavio\cite{DBLP:journals/corr/abs-1710-07557}} & 66\%  & 96\% \\
     \textbf{Uttara \cite{9299633}} & 67\% & 95\% \\
     \textbf{Akash  \cite{DBLP:journals/corr/abs-1910-05602}} & 60.58\% & ---\\
     \textbf{Mohammed  \cite{9378708}} & --- & 94.49\% \\
     \hline\noalign{\smallskip}
\end{tabular}
\end{table}

\begin{figure}[ht]
  \centering
  \includegraphics[width=\linewidth]{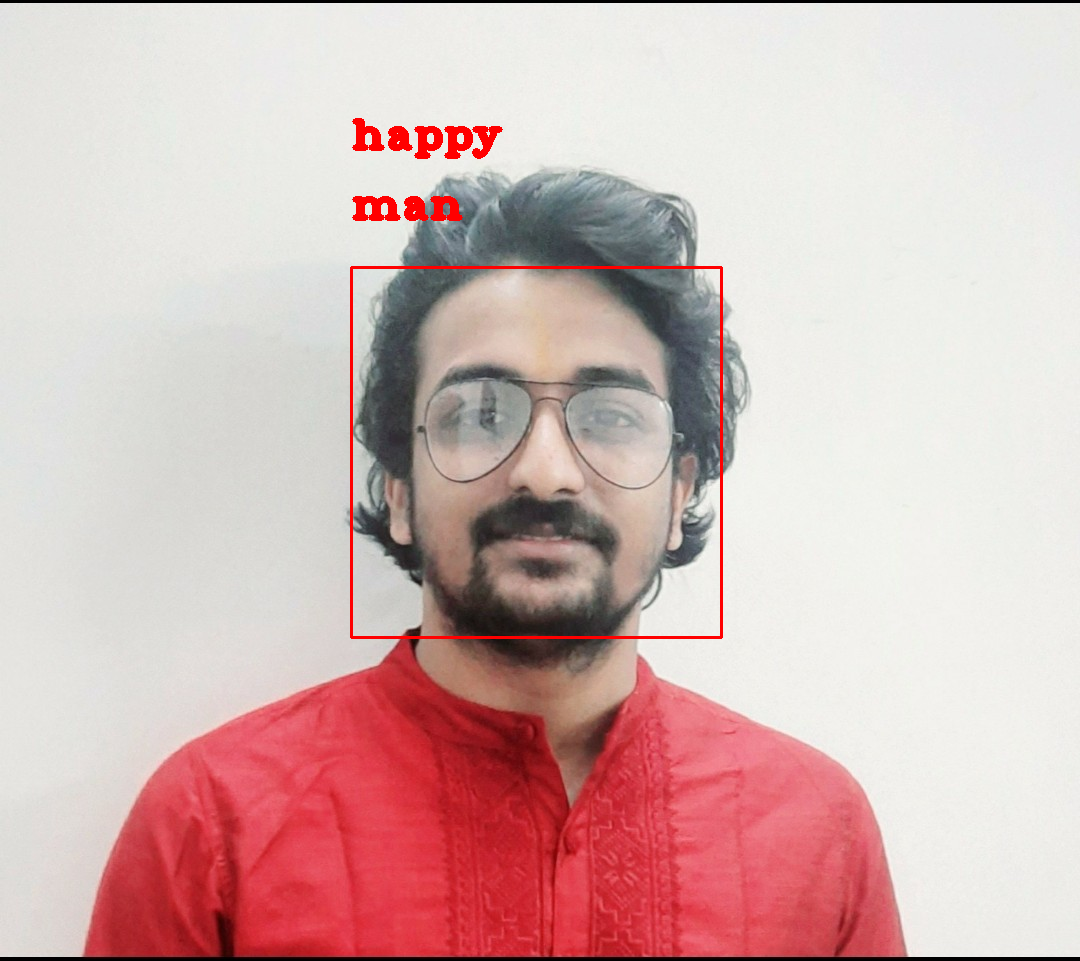}
  \caption{single face emotion and gender classification when image is passed as an input}
\end{figure}

\begin{figure}[ht]
  \centering
  \includegraphics[width=\linewidth]{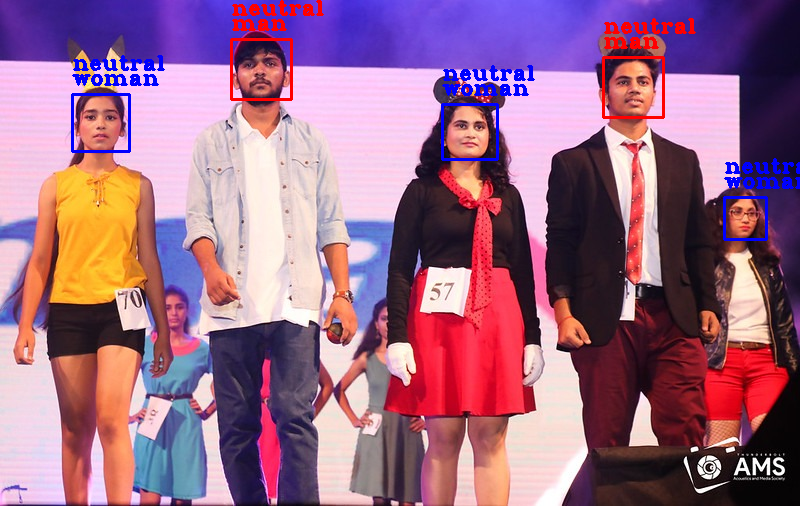}
  \caption{multiple faces emotion and gender classification 
  when image is passed as an input \cite{iiita_2019} . As it can be seen that our implementation can work on group images as well  with accurate results.}
\end{figure}

\begin{figure}[ht]
  \centering
  \includegraphics[width=\linewidth]{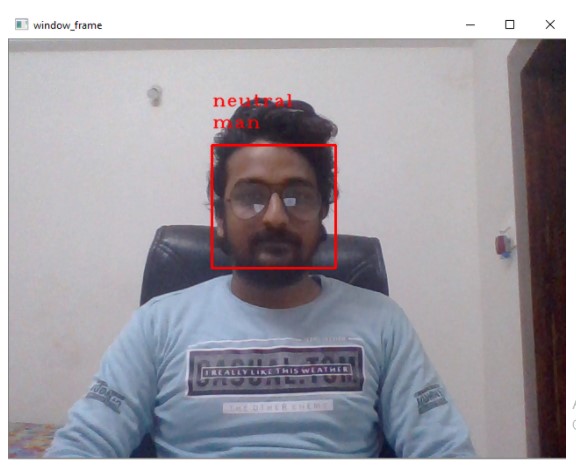}
  \caption{Webcam based Emotion and Gender Classification . Time for Predicting  Emotion and Gender of a person through webcam is 0.2 sec}
\end{figure}
\section{Conclusion}

Proposed model can be used for general classification purpose while retaining real-time inferences. Finally a real-time system is built that executes :face detection, emotion classification and gender classification into a single module.

This model can recognize the emotions by analysing the facial expression of the person shown to the webcam as an input to the model along with his/her gender. Emotion on the person’s face is classified into one of the seven categories-Angry , Disgust , Fear , Happy , Sad , Surprise , Neutral and then gender of the person is predicted as Male or Female.

\section{Future Work}
In emotion classification FER-2013, We can see that several common misclassifications like predicting “sad” instead of “fear” and predicting “angry” rather than  “disgust” . This is due to the label “angry” is more triggered when a person is glowering and these state features have mixed up among shadowy frames. While in gender classification trained CNNs are partial towards features and accessories of western artists. This misclassification is caused because the IMDB dataset consists of mostly western artists. We understand  that it is extremely important to uncover these behaviors  when using for real-time inferences.These can be overcomed by using different vast datasets for this purpose or by uniformly distributing images for each class.

\bibliographystyle{IEEEtran}
\bibliography{abc.bib}

\begin{thebibliography}{10}
\providecommand{\url}[1]{#1}
\csname url@samestyle\endcsname
\providecommand{\newblock}{\relax}
\providecommand{\bibinfo}[2]{#2}
\providecommand{\BIBentrySTDinterwordspacing}{\spaceskip=0pt\relax}
\providecommand{\BIBentryALTinterwordstretchfactor}{4}
\providecommand{\BIBentryALTinterwordspacing}{\spaceskip=\fontdimen2\font plus
\BIBentryALTinterwordstretchfactor\fontdimen3\font minus
  \fontdimen4\font\relax}
\providecommand{\BIBforeignlanguage}[2]{{%
\expandafter\ifx\csname l@#1\endcsname\relax
\typeout{** WARNING: IEEEtran.bst: No hyphenation pattern has been}%
\typeout{** loaded for the language `#1'. Using the pattern for}%
\typeout{** the default language instead.}%
\else
\language=\csname l@#1\endcsname
\fi
#2}}
\providecommand{\BIBdecl}{\relax}
\BIBdecl

\bibitem{DBLP:journals/corr/abs-1710-07557}
\BIBentryALTinterwordspacing
O.~Arriaga, M.~Valdenegro{-}Toro, and P.~Pl{\"{o}}ger, ``Real-time
  convolutional neural networks for emotion and gender classification,''
  \emph{CoRR}, vol. abs/1710.07557, 2017. [Online]. Available:
  \url{http://arxiv.org/abs/1710.07557}
\BIBentrySTDinterwordspacing

\bibitem{DBLP:journals/corr/Chollet16a}
\BIBentryALTinterwordspacing
F.~Chollet, ``Xception: Deep learning with depthwise separable convolutions,''
  \emph{CoRR}, vol. abs/1610.02357, 2016. [Online]. Available:
  \url{http://arxiv.org/abs/1610.02357}
\BIBentrySTDinterwordspacing

\bibitem{7780677}
C.~Szegedy, V.~Vanhoucke, S.~Ioffe, J.~Shlens, and Z.~Wojna, ``Rethinking the
  inception architecture for computer vision,'' in \emph{2016 IEEE Conference
  on Computer Vision and Pattern Recognition (CVPR)}, 2016, pp. 2818--2826.

\bibitem{9299633}
U.~Gogate, A.~Parate, S.~Sah, and S.~Narayanan, ``Real time emotion recognition
  and gender classification,'' in \emph{2020 International Conference on Smart
  Innovations in Design, Environment, Management, Planning and Computing
  (ICSIDEMPC)}, 2020, pp. 138--143.

\bibitem{DBLP:journals/corr/abs-1910-05602}
\BIBentryALTinterwordspacing
A.~Saravanan, G.~Perichetla, and K.~S. Gayathri, ``Facial emotion recognition
  using convolutional neural networks,'' \emph{CoRR}, vol. abs/1910.05602,
  2019. [Online]. Available: \url{http://arxiv.org/abs/1910.05602}
\BIBentrySTDinterwordspacing

\bibitem{DBLP:journals/corr/HowardZCKWWAA17}
\BIBentryALTinterwordspacing
A.~G. Howard, M.~Zhu, B.~Chen, D.~Kalenichenko, W.~Wang, T.~Weyand,
  M.~Andreetto, and H.~Adam, ``Mobilenets: Efficient convolutional neural
  networks for mobile vision applications,'' \emph{CoRR}, vol. abs/1704.04861,
  2017. [Online]. Available: \url{http://arxiv.org/abs/1704.04861}
\BIBentrySTDinterwordspacing

\bibitem{9378708}
M.~K. Benkaddour, S.~Lahlali, and M.~Trabelsi, ``Human age and gender
  classification using convolutional neural network,'' in \emph{2020 2nd
  International Workshop on Human-Centric Smart Environments for Health and
  Well-being (IHSH)}, 2021, pp. 215--220.

\bibitem{iiita_2019}
\BIBentryALTinterwordspacing
IIITA, \emph{IIITA}.\hskip 1em plus 0.5em minus 0.4em\relax IIITA, 2019,
  accessed 2021-06-26. [Online]. Available:
  \url{https://www.flickr.com/photos/ams_iiita/48930772538/in/\\album-72157711423840627/}
\BIBentrySTDinterwordspacing

\bibitem{8697761}
S.~M. Deokar, S.~S. Patankar, and J.~V. Kulkarni, ``Prominent face region based
  gender classification using deep learning,'' in \emph{2018 Fourth
  International Conference on Computing Communication Control and Automation
  (ICCUBEA)}, 2018, pp. 1--4.

\bibitem{8673352}
B.~Ramdhani, E.~C. Djamal, and R.~Ilyas, ``Convolutional neural networks models
  for facial expression recognition,'' in \emph{2018 International Symposium on
  Advanced Intelligent Informatics (SAIN)}, 2018, pp. 96--101.

\bibitem{7780459}
K.~He, X.~Zhang, S.~Ren, and J.~Sun, ``Deep residual learning for image
  recognition,'' in \emph{2016 IEEE Conference on Computer Vision and Pattern
  Recognition (CVPR)}, 2016, pp. 770--778.

\bibitem{10.1016/j.neunet.2014.09.005}
\BIBentryALTinterwordspacing
I.~J. Goodfellow, ``Challenges in representation learning,'' \emph{Neural
  Netw.}, vol.~64, no.~C, p. 59–63, Apr. 2015. [Online]. Available:
  \url{https://doi.org/10.1016/j.neunet.2014.09.005}
\BIBentrySTDinterwordspacing

\bibitem{6514409}
N.~U. Khan, ``A comparative analysis of facial expression recognition
  techniques,'' in \emph{2013 3rd IEEE International Advance Computing
  Conference (IACC)}, 2013, pp. 1262--1268.

\bibitem{7294547}
J.~Li and E.~Y. Lam, ``Facial expression recognition using deep neural
  networks,'' in \emph{2015 IEEE International Conference on Imaging Systems
  and Techniques (IST)}, 2015, pp. 1--6.

\bibitem{DBLP:journals/corr/IoffeS15}
\BIBentryALTinterwordspacing
S.~Ioffe and C.~Szegedy, ``Batch normalization: Accelerating deep network
  training by reducing internal covariate shift,'' \emph{CoRR}, vol.
  abs/1502.03167, 2015. [Online]. Available:
  \url{http://arxiv.org/abs/1502.03167}
\BIBentrySTDinterwordspacing

\bibitem{kingma2017adam}
D.~P. Kingma and J.~Ba, ``Adam: A method for stochastic optimization,'' 2017.

\bibitem{Mehendale2020}
\BIBentryALTinterwordspacing
N.~Mehendale, ``Facial emotion recognition using convolutional neural networks
  (ferc),'' \emph{SN Applied Sciences}, vol.~2, no.~3, p. 446, Feb 2020.
  [Online]. Available: \url{https://doi.org/10.1007/s42452-020-2234-1}
\BIBentrySTDinterwordspacing

\bibitem{chi-feng}
\BIBentryALTinterwordspacing
C.-F. Wang, ``A basic introduction to separable convolutions,'' Aug 2018,
  accessed = 26-06-2021. [Online]. Available:
  \url{https://towardsdatascience.com/a-basic-introduction-to-separable-convolutions-b99ec3102728}
\BIBentrySTDinterwordspacing

\bibitem{pandey_2018}
\BIBentryALTinterwordspacing
A.~Pandey, ``Depth-wise convolution and depth-wise separable convolution,'' Sep
  2018, accessed = 25-06-2021. [Online]. Available:
  \url{https://medium.com/@zurister/depth-wise-convolution-and-depth-wise-separable-convolution-37346565d4ec}
\BIBentrySTDinterwordspacing

\end{thebibliography}
\nocite{8697761}
\nocite{8673352}
\nocite{7780459}
\nocite{10.1016/j.neunet.2014.09.005}
\nocite{9378708}
\nocite{6514409}
\nocite{7294547}
\nocite{DBLP:journals/corr/IoffeS15}
\nocite{kingma2017adam}
\nocite{Mehendale2020}
\nocite{chi-feng}
\nocite{pandey_2018}

\end{document}